\documentclass{article}
\PassOptionsToPackage{numbers,sort&compress}{natbib}
\usepackage[dblblindworkshop,nonanonymous]{neurips_2026}
\AtBeginDocument{\nolinenumbers}
\providecommand{\workshoptitle}[1]{}
\workshoptitle{Agentic AI for Biological Discovery: Toward Closed-Loop Life-Science Intelligence}
\makeatletter
\renewcommand{\@noticestring}{%
  \@neuripsordinal\ Conference on Neural Information Processing Systems (NeurIPS \@neuripsyear)
  Workshop: \@workshoptitle.%
}
\makeatother

\usepackage[utf8]{inputenc}
\usepackage[T1]{fontenc}
\usepackage{hyperref}
\usepackage{url}
\usepackage{booktabs,longtable,array,multirow}
\usepackage{amsfonts,amsmath}
\usepackage{microtype}
\usepackage{xcolor}
\usepackage{graphicx}
\usepackage{pifont}
\usepackage{enumitem}
\usepackage{listings}
\usepackage{wrapfig}

\graphicspath{{assets/figures/}}
\setlist{nosep,leftmargin=*}
\newcolumntype{P}[1]{>{\raggedright\arraybackslash}p{#1}}

\newcommand{\pmark}{\textcolor{gray!80!black}{$\circ$}}
\newcommand{\ok}{\textcolor{green!55!black}{\ding{51}}}
\newcommand{\no}{\textcolor{red!75!black}{\ding{55}}}

\newcommand{\bench}{BioStudyBench}

\makeatletter\newcommand{\R}[2]{\@ifundefined{R@#1@#2}{\textbf{??}}{\csname R@#1@#2\endcsname}}\makeatother
\expandafter\def\csname R@claude-opus-5-5@valid\endcsname{23}
\expandafter\def\csname R@claude-opus-5-5@loose\endcsname{23}
\expandafter\def\csname R@claude-opus-5-5@strict\endcsname{23}
\expandafter\def\csname R@claude-opus-5-5@tight\endcsname{22}
\expandafter\def\csname R@claude-opus-5-5@paired\endcsname{23}
\expandafter\def\csname R@claude-opus-5-5@pairedstrict\endcsname{23}
\expandafter\def\csname R@claude-opus-5-5@ndstrict\endcsname{10}
\expandafter\def\csname R@claude-opus-5-5@lift\endcsname{13}
\expandafter\def\csname R@claude-opus-5-5@unleaked\endcsname{13}
\expandafter\def\csname R@claude-opus-5-5@unleakedn\endcsname{13}
\expandafter\def\csname R@claude-opus-5-5@infra\endcsname{0}
\expandafter\def\csname R@claude-opus-5-5@cost\endcsname{5.44}
\expandafter\def\csname R@claude-opus-5-5@wall\endcsname{28}
\expandafter\def\csname R@claude-opus-5-5@tokens\endcsname{8.2}
\expandafter\def\csname R@gpt-6-astra@valid\endcsname{24}
\expandafter\def\csname R@gpt-6-astra@loose\endcsname{23}
\expandafter\def\csname R@gpt-6-astra@strict\endcsname{23}
\expandafter\def\csname R@gpt-6-astra@tight\endcsname{23}
\expandafter\def\csname R@gpt-6-astra@paired\endcsname{24}
\expandafter\def\csname R@gpt-6-astra@pairedstrict\endcsname{23}
\expandafter\def\csname R@gpt-6-astra@ndstrict\endcsname{15}
\expandafter\def\csname R@gpt-6-astra@lift\endcsname{8}
\expandafter\def\csname R@gpt-6-astra@unleaked\endcsname{9}
\expandafter\def\csname R@gpt-6-astra@unleakedn\endcsname{9}
\expandafter\def\csname R@gpt-6-astra@infra\endcsname{0}
\expandafter\def\csname R@gpt-6-astra@cost\endcsname{6.44}
\expandafter\def\csname R@gpt-6-astra@wall\endcsname{18}
\expandafter\def\csname R@gpt-6-astra@tokens\endcsname{3.1}
\expandafter\def\csname R@gpt-6-luna@valid\endcsname{24}
\expandafter\def\csname R@gpt-6-luna@loose\endcsname{20}
\expandafter\def\csname R@gpt-6-luna@strict\endcsname{19}
\expandafter\def\csname R@gpt-6-luna@tight\endcsname{19}
\expandafter\def\csname R@gpt-6-luna@paired\endcsname{24}
\expandafter\def\csname R@gpt-6-luna@pairedstrict\endcsname{19}
\expandafter\def\csname R@gpt-6-luna@ndstrict\endcsname{5}
\expandafter\def\csname R@gpt-6-luna@lift\endcsname{14}
\expandafter\def\csname R@gpt-6-luna@unleaked\endcsname{14}
\expandafter\def\csname R@gpt-6-luna@unleakedn\endcsname{19}
\expandafter\def\csname R@gpt-6-luna@infra\endcsname{0}
\expandafter\def\csname R@gpt-6-luna@cost\endcsname{0.16}
\expandafter\def\csname R@gpt-6-luna@wall\endcsname{13}
\expandafter\def\csname R@gpt-6-luna@tokens\endcsname{9.1}
\expandafter\def\csname R@gemini-3.8-flash@valid\endcsname{25}
\expandafter\def\csname R@gemini-3.8-flash@loose\endcsname{24}
\expandafter\def\csname R@gemini-3.8-flash@strict\endcsname{24}
\expandafter\def\csname R@gemini-3.8-flash@tight\endcsname{24}
\expandafter\def\csname R@gemini-3.8-flash@paired\endcsname{25}
\expandafter\def\csname R@gemini-3.8-flash@pairedstrict\endcsname{24}
\expandafter\def\csname R@gemini-3.8-flash@ndstrict\endcsname{11}
\expandafter\def\csname R@gemini-3.8-flash@lift\endcsname{13}
\expandafter\def\csname R@gemini-3.8-flash@unleaked\endcsname{13}
\expandafter\def\csname R@gemini-3.8-flash@unleakedn\endcsname{14}
\expandafter\def\csname R@gemini-3.8-flash@infra\endcsname{0}
\expandafter\def\csname R@gemini-3.8-flash@cost\endcsname{1.62}
\expandafter\def\csname R@gemini-3.8-flash@wall\endcsname{14}
\expandafter\def\csname R@gemini-3.8-flash@tokens\endcsname{11.5}
\expandafter\def\csname R@glm-5.3@valid\endcsname{23}
\expandafter\def\csname R@glm-5.3@loose\endcsname{19}
\expandafter\def\csname R@glm-5.3@strict\endcsname{19}
\expandafter\def\csname R@glm-5.3@tight\endcsname{18}
\expandafter\def\csname R@glm-5.3@paired\endcsname{23}
\expandafter\def\csname R@glm-5.3@pairedstrict\endcsname{19}
\expandafter\def\csname R@glm-5.3@ndstrict\endcsname{7}
\expandafter\def\csname R@glm-5.3@lift\endcsname{12}
\expandafter\def\csname R@glm-5.3@unleaked\endcsname{12}
\expandafter\def\csname R@glm-5.3@unleakedn\endcsname{16}
\expandafter\def\csname R@glm-5.3@infra\endcsname{0}
\expandafter\def\csname R@glm-5.3@cost\endcsname{1.73}
\expandafter\def\csname R@glm-5.3@wall\endcsname{11}
\expandafter\def\csname R@glm-5.3@tokens\endcsname{4.9}
\expandafter\def\csname R@mimo-v2.6-pro@valid\endcsname{21}
\expandafter\def\csname R@mimo-v2.6-pro@loose\endcsname{19}
\expandafter\def\csname R@mimo-v2.6-pro@strict\endcsname{19}
\expandafter\def\csname R@mimo-v2.6-pro@tight\endcsname{19}
\expandafter\def\csname R@mimo-v2.6-pro@paired\endcsname{21}
\expandafter\def\csname R@mimo-v2.6-pro@pairedstrict\endcsname{19}
\expandafter\def\csname R@mimo-v2.6-pro@ndstrict\endcsname{5}
\expandafter\def\csname R@mimo-v2.6-pro@lift\endcsname{14}
\expandafter\def\csname R@mimo-v2.6-pro@unleaked\endcsname{14}
\expandafter\def\csname R@mimo-v2.6-pro@unleakedn\endcsname{16}
\expandafter\def\csname R@mimo-v2.6-pro@infra\endcsname{0}
\expandafter\def\csname R@mimo-v2.6-pro@cost\endcsname{0.19}
\expandafter\def\csname R@mimo-v2.6-pro@wall\endcsname{48}
\expandafter\def\csname R@mimo-v2.6-pro@tokens\endcsname{3.7}
\expandafter\def\csname R@deepseek-v4.1-flash@valid\endcsname{21}
\expandafter\def\csname R@deepseek-v4.1-flash@loose\endcsname{19}
\expandafter\def\csname R@deepseek-v4.1-flash@strict\endcsname{19}
\expandafter\def\csname R@deepseek-v4.1-flash@tight\endcsname{19}
\expandafter\def\csname R@deepseek-v4.1-flash@paired\endcsname{21}
\expandafter\def\csname R@deepseek-v4.1-flash@pairedstrict\endcsname{19}
\expandafter\def\csname R@deepseek-v4.1-flash@ndstrict\endcsname{8}
\expandafter\def\csname R@deepseek-v4.1-flash@lift\endcsname{11}
\expandafter\def\csname R@deepseek-v4.1-flash@unleaked\endcsname{12}
\expandafter\def\csname R@deepseek-v4.1-flash@unleakedn\endcsname{13}
\expandafter\def\csname R@deepseek-v4.1-flash@infra\endcsname{0}
\expandafter\def\csname R@deepseek-v4.1-flash@cost\endcsname{0.06}
\expandafter\def\csname R@deepseek-v4.1-flash@wall\endcsname{57}
\expandafter\def\csname R@deepseek-v4.1-flash@tokens\endcsname{11.8}
\expandafter\def\csname R@qwen3.8-27b@valid\endcsname{18}
\expandafter\def\csname R@qwen3.8-27b@loose\endcsname{16}
\expandafter\def\csname R@qwen3.8-27b@strict\endcsname{16}
\expandafter\def\csname R@qwen3.8-27b@tight\endcsname{16}
\expandafter\def\csname R@qwen3.8-27b@paired\endcsname{18}
\expandafter\def\csname R@qwen3.8-27b@pairedstrict\endcsname{16}
\expandafter\def\csname R@qwen3.8-27b@ndstrict\endcsname{3}
\expandafter\def\csname R@qwen3.8-27b@lift\endcsname{13}
\expandafter\def\csname R@qwen3.8-27b@unleaked\endcsname{13}
\expandafter\def\csname R@qwen3.8-27b@unleakedn\endcsname{15}
\expandafter\def\csname R@qwen3.8-27b@infra\endcsname{0}
\expandafter\def\csname R@qwen3.8-27b@cost\endcsname{1.72}
\expandafter\def\csname R@qwen3.8-27b@wall\endcsname{69}
\expandafter\def\csname R@qwen3.8-27b@tokens\endcsname{12.3}

\newcommand{\blindNonnullPct}{83}

\newcommand{\blindNullPct}{16}

\newcommand{\blindN}{2{,}131}
\newcommand{\blindSigAllPct}{80}
\newcommand{\blindSigNullK}{63}
\newcommand{\blindSigNullN}{104}
\newcommand{\blindSigNullPct}{61}
\newcommand{\blindUnknownPct}{85}

\newcommand{\ppFixedMin}{69}
\newcommand{\ppFixedMax}{84}

\newcommand{\snkHarvest}{404{,}019}

\newcommand{\snkShortlist}{2{,}131}
\newcommand{\snkGood}{35}
\newcommand{\snkAuthored}{33}
\newcommand{\snkOpen}{25}

\newcommand{\seedNdMin}{24}
\newcommand{\seedNdMax}{51}

\newcommand{\seedOpenItems}{25}

\newcommand{\seedGainMean}{47}
\newcommand{\seedClosedNd}{40}
\newcommand{\seedOpenNd}{26}
\newcommand{\seedClosedOp}{89}
\newcommand{\seedOpenOp}{70}

\newcommand{\seedGainMin}{32}
\newcommand{\seedGainMax}{55}
\newcommand{\seedBestClosedOp}{94.7}
\newcommand{\seedBestOpenOp}{81.3}

\newcommand{\seedTurnsMax}{117}

\newcommand{\predKnownNoRate}{81}
\newcommand{\predUnsureRate}{95}
\newcommand{\predMonthMin}{82}
\newcommand{\predMonthMax}{85}

\newcommand{\leakToolCalls}{17{,}429}

\newcommand{\leakRecordsReturned}{8{,}071}
\newcommand{\leakRecordsAfterCutoff}{0}

\title{\bench: Evaluating Agents on Post-Cutoff Biomedical Studies}
\author{%
  David Li \\
  University of Waterloo \\
  \href{mailto:david.li3@uwaterloo.ca}{\texttt{david.li3@uwaterloo.ca}}
  \and
  Shaamil Karim \\
  Atlas Discovery \\
  \href{mailto:shaamil@atlasdiscovery.bio}{\texttt{shaamil@atlasdiscovery.bio}}
  \and
  Christian Gensbigler \\
  Atlas Discovery \\
  \href{mailto:christian@atlasdiscovery.bio}{\texttt{christian@atlasdiscovery.bio}}
}
\makeatletter
\renewcommand{\@author}{%
  David Li\textsuperscript{\textmd{1}},
  Shaamil Karim\textsuperscript{\textmd{2}},
  Christian Gensbigler\textsuperscript{\textmd{2}} \\
  \textsuperscript{1}University of Waterloo\quad
  \textsuperscript{2}Atlas Discovery \\
  {\small\texttt{david.li3@uwaterloo.ca}} \\
  {\small\texttt{\{shaamil, christian\}@atlasdiscovery.bio}}
}
\makeatother

\begin{document}
\maketitle

\begin{abstract}
We evaluate whether AI agents can match the reported findings of published biomedical studies using public data. Existing evaluations do not consistently separate analysis from prior knowledge or retrieval of the published answer. We introduce \bench{}, a benchmark of \snkOpen{} long-horizon analysis tasks drawn from studies first published between July and September 2026, after the developer-reported knowledge cutoffs of the models we evaluate, semi-automatically filtered down from \snkHarvest{} PubMed records. In each task, the agent receives a neutral research question but no data files, so it must find and download the relevant public data, search the literature through tools that return only records dated before its cutoff, and report findings through data analysis. To measure gains over prior knowledge, we run every task both with and without access to data and tools. Across eight models, access to data and tools raises the pass rate by \seedGainMean{} percentage points on average over the no-data baseline. Open-weight models across sizes trail closed-weight models, with the best open-weight model passing \seedBestOpenOp\% of tasks against \seedBestClosedOp\% for the best closed-weight model.
\end{abstract}

\section{Introduction}

Re-deriving a published biomedical finding requires locating the right data release, reconstructing the study population, applying the survey or study design and choosing an estimator. Current agents do this unreliably. The best agents reported in the respective papers answer 21\% of BixBench's open-ended analysis questions \citep{bixbench}, succeed on 39.9\% of RWE-bench's observational-study tasks \citep{rwebench}, and achieve a mean artifact pass rate across tasks of 48\% on BixBench3 \citep{bixbench3}.

These scores are also difficult to interpret (Table~\ref{tab:compare}). These data-analysis benchmarks do not require every source study to postdate every evaluated model's cutoff, so a correct answer alone does not establish that a finding was newly derived. PaperBench and FIRE-Bench allow web access without date restrictions \citep{paperbench,firebench}, and \citet{sciconbench} report that agents retrieved the ground-truth conclusion ``even when instructed not to''. We found that given only the research question for each of \blindN{} biomedical papers published after its cutoff, GPT-6 Sol reported no knowledge of \blindUnknownPct\% of the findings, yet matched the direction and significance of \blindNonnullPct\% of the non-null ones (Appendix~\ref{app:predict}).

\begin{figure}[t]
\centering
\includegraphics[width=\linewidth]{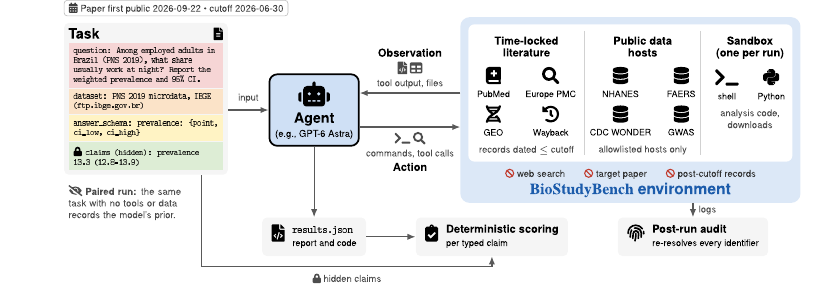}
\caption{\textbf{Overview of the \bench{} environment}, shown for one task. The agent sees the question (red), dataset (orange) and answer schema (yellow) but not the reference claims (green). Its tools return only records dated before its cutoff, and its answers are scored against the hidden findings; an audit checks the logs, and a paired run without data records the model's prior.}
\label{fig:overview}
\end{figure}

We introduce \bench{}, an environment and benchmark for evaluating whether agents can match findings from studies published after their developer-reported knowledge cutoff (Figure~\ref{fig:overview}). The agent is given a research question but not the data. It must find and download relevant public data, write scripts to perform analysis, and report its findings. Each task requires long-horizon analysis, taking agents up to \seedTurnsMax{} turns per task on average, with runs lasting up to 90 minutes.

\begin{table}[t]
\centering\small
\caption{\textbf{Selected benchmarks for scientific data analysis and conclusion synthesis.} \ok{} yes; \pmark{} partial, or as an ablation only; \no{} no; -- not applicable. Post-cutoff means a required study-date split, rather than possible inclusion of recent studies.}
\label{tab:compare}
{\fontsize{7.3}{8.8}\selectfont
\setlength{\tabcolsep}{3.25pt}
\begin{tabular}{@{}llcccccc@{}}
\toprule
 & & & Post-cutoff & No-data & Date-restricted & Agent acquires & Deterministic \\
 & \# Tasks & Domain & tasks? & control? & search? & the data? & scoring? \\
\midrule
\multicolumn{8}{@{}l}{\textit{Data analysis}} \\
DiscoveryBench \citep{discoverybench} & 264 & multiple & \no & \pmark & \no & \no & \no \\
BixBench \citep{bixbench} & 205 (61) & bioinformatics & \no & \pmark & \no & \no & \no \\
BixBench3 \citep{bixbench3} & 20 & computational biology & \no & \no & \no & \no & \ok \\
RWE-bench \citep{rwebench} & 162 & EHR (MIMIC-IV) & \no & \no & \no & \no & \ok \\
MedResearchBench \citep{medresearchbench} & 16 & NHANES, SEER & \no & \no & \no & \pmark & \no \\
\midrule
\multicolumn{8}{@{}l}{\textit{Time-controlled evaluation}} \\
FIRE-Bench \citep{firebench} & 40 & machine learning & \pmark & \no & \no & \no & \no \\
SciConBench \citep{sciconbench} & 268 & systematic reviews & \ok & \pmark & \ok & -- & \no \\
\midrule
\bench{} & \snkOpen{} & biomedical studies & \ok & \ok & \ok & \ok & \ok \\
\bottomrule
\end{tabular}}
\par\smallskip
{\footnotesize\raggedright BixBench: 205 questions in 61 capsules (v3). MedResearchBench: 16 tasks, with pilot results on 3. FIRE-Bench: 40 evaluated tasks, plus 60 additional released papers. SciConBench: 268 post-cutoff reviews selected from 9,107; search is restricted by the target review's publication date, and no-tool controls cover three base models.\par}
\end{table}

\section{Related work}\label{sec:related}
\paragraph{Automated data analysis.}
DiscoveryBench \citep{discoverybench} derives data-driven hypotheses from papers in six fields and scores them with a language-model judge; BixBench \citep{bixbench} poses open-answer bioinformatics questions, and BixBench3 \citep{bixbench3} asks agents to reproduce artifacts from computational-biology studies given each study's protocol. Closest to our setting, RWE-bench \citep{rwebench} evaluates agents on 162 observational-study tasks in MIMIC-IV, while MedResearchBench \citep{medresearchbench} offers 16 tasks using NHANES and SEER, with pilot results on three. Other benchmarks give the agent a paper's code or text to reproduce \citep{corebench,paperbench,replicationbench} or pose data-science questions on supplied data \citep{blade,scienceagentbench,biodsa}. Our emphasis is on combining post-cutoff studies, restricted retrieval, and paired runs with and without data.

\paragraph{Information control.}
Live benchmarks refresh questions using recent sources or newly emerging knowledge to limit contamination \citep{livebench,antileak}. FIRE-Bench \citep{firebench} asks agents to rediscover the insights of 40 machine-learning papers, including both pre- and post-cutoff studies, with unrestricted web access. SciConBench \citep{sciconbench}, closest to our leakage controls, filters search results by the target review's publication date and blocks reference-review content, but asks agents to synthesise review conclusions rather than analyse data; Proof of Time \citep{proofoftime} freezes evidence to forecast post-cutoff outcomes. Language models can distinguish reported neuroscience results from altered alternatives \citep{brainbench}, and analysts diverge on the same data \citep{Silberzahn2018ManyAO}, motivating our no-data control and tolerances. See Table~\ref{tab:compare} for a comparison with these benchmarks.

\section{The \bench{} environment}\label{sec:env}
\paragraph{Tasks.}
We source tasks in \bench{} from published biomedical papers. Our semi-automated pipeline (Section~\ref{sec:bench}) turns each paper into a neutral question and uses the paper's findings as reference answers. The agent is not given the data. The task names the public source, and the agent must download the raw files, read the codebooks, and merge, recode and weight the records itself. It reports its answers in a fixed format, and a rule-based scorer compares them with the paper's findings. A task passes if its primary estimate falls within the original paper's 95\% confidence interval, or, for a finding reported only as an association, if its direction and statistical significance match the paper's. Our score measures agreement with the study's reported findings. The reference confidence interval provides an operational scoring tolerance; a passing answer does not by itself validate the agent's data preparation, statistical methods, uncertainty estimates, or scientific interpretation. The agent works in a sandbox with a shell and Python and searches the literature and datasets through tools covering PubMed, Europe PMC, OpenAlex, Crossref, GEO, ClinicalTrials.gov, Zenodo and the Wayback Machine.

\paragraph{Leakage controls.}\label{sec:controls}
Every task in \bench{} originates from a paper first published between July and September 2026, after June 2026, the latest developer-reported knowledge cutoff among the evaluated models (Appendix Table~\ref{tab:cutoffs}). This temporal separation, together with target-paper blocklists and restricted retrieval, reduces contamination risk. However, published cutoffs and publication dates do not establish the absence of prior exposure through all training or post-training data. Besides the literature and dataset tools listed above, the agent's shell can reach only allowlisted public data hosts, such as the CDC and FDA servers that distribute NHANES and FAERS, and web search is disabled (Figure~\ref{fig:overview}). The literature and dataset tools filter returned records by their reported dates and apply target-paper blocklists (Appendix~\ref{app:controls}).

\begin{figure}[t]
\centering
\includegraphics[width=\linewidth]{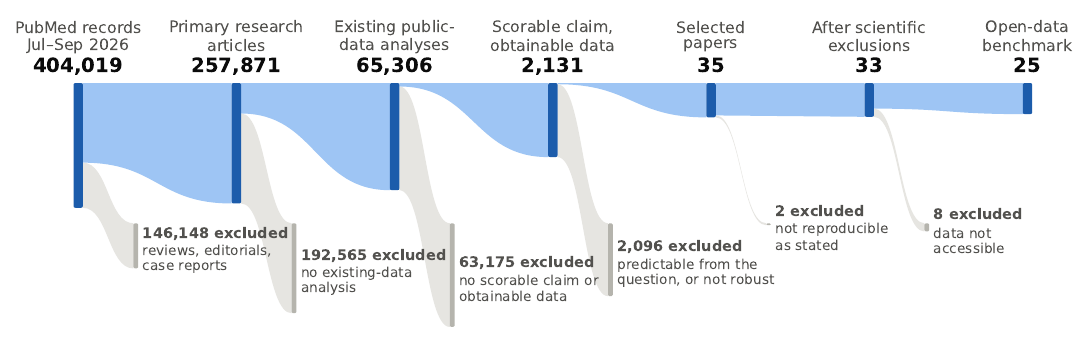}
\caption{\textbf{Construction funnel.} Node heights are log-scaled; each node splits in proportion between kept (blue) and excluded (grey) records.}
\label{fig:sankey}
\end{figure}

\begin{figure}[t]
\centering
\includegraphics[width=\linewidth]{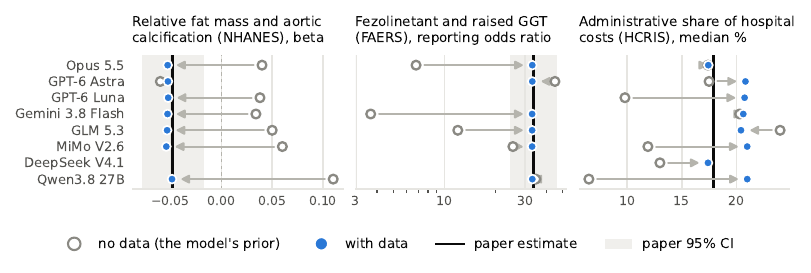}
\caption{\textbf{Answers without data (hollow) and with data (blue)} on two tasks where models match the reference with data (left, centre) and one they struggle with (right).}
\label{fig:examples}
\end{figure}

\section{Benchmark construction and experimental setup}\label{sec:bench}
As shown in Figure~\ref{fig:sankey}, we retained the \snkHarvest{} PubMed \citep{sayers2026ncbi} records first made public between July and September 2026 and resolved each record's earliest appearance across preprint servers and publishers. We designed a structural filter to remove reviews, editorials and case reports, and two classifier passes (GPT-6 Luna) which retain \snkShortlist{} studies of existing public data with a single scorable claim and obtainable data. A \emph{recall screen} then showed GPT-6 Sol only each question, demoted findings it predicted, and a review classifier kept \snkGood{} studies (full prompts provided in Appendix~\ref{app:prompts}). After scientific exclusions, \snkAuthored{} tasks remain; \snkOpen{} have accessible data and form BioStudyBench (Appendix~\ref{app:items}).

We run each model on every task under two conditions, three times each. In the \emph{with data and tools} condition, the agent works in the full environment, with limits of 300 turns and 90 minutes. In the \emph{prior knowledge} condition, it receives the same question and answer format but no tools or data, and must predict every requested value from what it already knows. Each repeat's pass rate uses all \seedOpenItems{} tasks as its denominator; timeouts, run errors and runs without scored answers count as failures.

\section{Results}\label{sec:results}

\begin{wraptable}[16]{r}{0.47\linewidth}
\vspace{-12pt}
\centering\footnotesize
\setlength{\tabcolsep}{3pt}
\begin{tabular}{@{}lrrr@{}}
\toprule
 & \multicolumn{2}{c}{Pass rate (\%)} & \\
\cmidrule(lr){2-3}
Model & \begin{tabular}[b]{@{}c@{}}Prior\\knowledge\end{tabular} & \begin{tabular}[b]{@{}c@{}}With data\\and tools\end{tabular} & Turns \\
\midrule
\multicolumn{4}{@{}l}{\textit{Closed weights}} \\
GPT-6 Astra & 50.7$_{\pm 8.3}$ & 94.7$_{\pm 2.3}$ & 42 \\
Gemini 3.8 Flash & 40.0$_{\pm 10.6}$ & 94.7$_{\pm 6.1}$ & 117 \\
Claude Opus 5.5 & 46.7$_{\pm 6.1}$ & 93.3$_{\pm 2.3}$ & 85 \\
GPT-6 Luna & 24.0$_{\pm 4.0}$ & 74.7$_{\pm 6.1}$ & 64 \\
\midrule
\multicolumn{4}{@{}l}{\textit{Open weights}} \\
GLM 5.3 & 26.7$_{\pm 2.3}$ & 81.3$_{\pm 6.1}$ & 65 \\
DeepSeek V4.1 Flash & 25.3$_{\pm 6.1}$ & 76.0$_{\pm 4.0}$ & 81 \\
MiMo V2.6 Pro & 26.7$_{\pm 6.1}$ & 66.7$_{\pm 12.9}$ & 43 \\
Qwen3.8 27B & 24.0$_{\pm 0.0}$ & 56.0$_{\pm 10.6}$ & 97 \\
\bottomrule
\end{tabular}

\caption{\textbf{Main results}: pass rate (\%) over \seedOpenItems{} tasks, mean $\pm$ s.d.\ across three repeats, and mean turns per run with data where recorded.}
\label{tab:results}
\vspace{-8pt}
\end{wraptable}
\paragraph{Data and tools improve reference agreement.}
Without data, models pass \seedNdMin--\seedNdMax\% of tasks; access to data and tools raises pass rates by \seedGainMin--\seedGainMax{} percentage points (Table~\ref{tab:results}). These gains do not establish that every successful answer was independently derived.

\paragraph{Open-weight models trail closed-weight models.}
Open-weight models average \seedOpenNd\% against \seedClosedNd\% for closed models without data, and \seedOpenOp\% against \seedClosedOp\% with data (Table~\ref{tab:results}). The best open-weight model, GLM~5.3, passes fewer tasks than three of the four closed-weight models. Figure~\ref{fig:examples} shows two tasks where widely differing priors converge on the published value once models analyse the data, and one on which they narrow but stay above it.

\paragraph{Qualitative failure cases.}
The inspected failures illustrate mismatches in analysis and reporting (Appendix Figure~\ref{fig:fail}). These failures span a range of modes: when a paper leaves an analytic choice implicit, models may choose an analysis different from the paper's; models may compute an estimate matching the reference but misstate the conclusion; and when a data source fails, models often fail to look for another route to the data.

\section{Limitations}
\bench{} contains only \snkOpen{} long-horizon tasks. Its temporal separation expires once evaluated models' reported cutoffs extend past September 2026. Our construction pipeline outlasts any one set of tasks, however, and can build a new window of post-cutoff papers for each generation of models.

Date filtering and host restrictions do not guarantee that every accessible source is free of answer-bearing information. Additionally, an incorrect analysis can produce a correct result, while a scientifically defensible alternative analysis can fall outside the scoring tolerance. Our reported pass rates therefore measure reference agreement rather than comprehensive methodological correctness.

\section{Conclusion}
\bench{} evaluates reference agreement on post-cutoff biomedical studies using restricted retrieval and paired no-data controls. Access to data and tools raises pass rates by \seedGainMin--\seedGainMax{} percentage points over prior knowledge alone; open-weight models trail closed-weight models in both conditions.

\bibliographystyle{plainnat}
\bibliography{assets/refs_all}

\begin{thebibliography}{23}
\providecommand{\natexlab}[1]{#1}
\providecommand{\url}[1]{\texttt{#1}}
\expandafter\ifx\csname urlstyle\endcsname\relax
  \providecommand{\doi}[1]{doi: #1}\else
  \providecommand{\doi}{doi: \begingroup \urlstyle{rm}\Url}\fi

\bibitem[{Anthropic}(2026)]{anthropic2026evalawareness}
{Anthropic}.
\newblock Eval awareness in claude opus 4.6's browsecomp performance.
\newblock
  \url{https://www.anthropic.com/engineering/eval-awareness-browsecomp}, March
  2026.
\newblock Accessed: 2026-09-27.

\bibitem[Chen et~al.(2024)Chen, Chen, Ning, Zhang, Wang, Yu, Li, Liao, Wei, Lu,
  Dey, Xue, Baker, Burns, Adu-Ampratwum, Huang, Ning, Gao, Su, and
  Sun]{scienceagentbench}
Ziru Chen, Shijie Chen, Yuting Ning, Qianheng Zhang, Boshi Wang, Botao Yu,
  Yifei Li, Zeyi Liao, Chen Wei, Zitong Lu, Vishal Dey, Mingyi Xue, Frazier~N.
  Baker, Benjamin Burns, Daniel Adu-Ampratwum, Xuhui Huang, Xia Ning, Song Gao,
  Yu~Su, and Huan Sun.
\newblock {ScienceAgentBench: Toward Rigorous Assessment of Language Agents for
  Data-Driven Scientific Discovery}.
\newblock \emph{arXiv preprint arXiv:2410.05080}, 2024.
\newblock URL \url{https://arxiv.org/abs/2410.05080}.
\newblock ICLR 2025.

\bibitem[Gao et~al.(2025)Gao, Zhang, Du, Mahindre, Somayajula, and
  Xie]{rewindtime}
Xin Gao, Ruiyi Zhang, Daniel Du, Saurabh Mahindre, Sai~Ashish Somayajula, and
  Pengtao Xie.
\newblock Can prompts rewind time for {LLMs}? evaluating the effectiveness of
  prompted knowledge cutoffs.
\newblock \emph{arXiv preprint arXiv:2510.02340}, 2025.
\newblock URL \url{https://arxiv.org/abs/2510.02340}.

\bibitem[Gu et~al.(2024)Gu, Shang, Jiang, Kuang, Lin, Lyu, Mao, Pan, Wu, Yu,
  Zhang, Zhang, Zhu, Merrill, Heer, and Althoff]{blade}
Ken Gu, Ruoxi Shang, Ruien Jiang, Keying Kuang, Richard-John Lin, Donghe Lyu,
  Yue Mao, Youran Pan, Teng Wu, Jiaqian Yu, Yikun Zhang, Tianmai~M. Zhang,
  Lanyi Zhu, Mike~A. Merrill, Jeffrey Heer, and Tim Althoff.
\newblock {BLADE: Benchmarking Language Model Agents for Data-Driven Science}.
\newblock \emph{arXiv preprint arXiv:2408.09667}, 2024.
\newblock URL \url{https://arxiv.org/abs/2408.09667}.

\bibitem[Jung et~al.(2026)Jung, Diniz, Roveda, da~Silva, Jung, Tsai, Korolova,
  and Ribeiro]{sciconbench}
Hayoung Jung, Pedro~Viana Diniz, Jos{\'e} Reinaldo~Corr{\^e}a Roveda,
  Abner~Fernandes da~Silva, Haeun Jung, Enoch Tsai, Aleksandra Korolova, and
  Manoel~Horta Ribeiro.
\newblock Can {AI} agents synthesize scientific conclusions?
\newblock \emph{arXiv preprint arXiv:2606.11337}, 2026.
\newblock URL \url{https://arxiv.org/abs/2606.11337}.

\bibitem[Koch et~al.(2026)Koch, Wassie, Valdes-Aleman, Lee, Hinks, Rodriques,
  White, and Laurent]{bixbench3}
Zane Koch, Asmamaw~T. Wassie, Javier Valdes-Aleman, Jason Lee, Michaela~M.
  Hinks, Samuel~G. Rodriques, Andrew~D. White, and Jon~M. Laurent.
\newblock {BixBench3: Benchmarking AI agents on research-study-scale
  computational biology tasks}.
\newblock \emph{arXiv preprint arXiv:2608.25286}, 2026.
\newblock URL \url{https://arxiv.org/abs/2608.25286}.

\bibitem[Li et~al.(2026)Li, He, Hu, Tian, and Li]{rwebench}
Dubai Li, Yuxiang He, Yan Hu, Yu~Tian, and Jingsong Li.
\newblock Can {LLM} agents generate real-world evidence? evaluating
  observational studies in medical databases.
\newblock \emph{arXiv preprint arXiv:2603.22767}, 2026.
\newblock URL \url{https://arxiv.org/abs/2603.22767}.

\bibitem[Lopez-Lira et~al.(2025)Lopez-Lira, Tang, and Zhu]{memorizationecon}
Alejandro Lopez-Lira, Yuehua Tang, and Mingyin Zhu.
\newblock The memorization problem: Can we trust {LLMs}' economic forecasts?
\newblock \emph{arXiv preprint arXiv:2504.14765}, 2025.
\newblock URL \url{https://arxiv.org/abs/2504.14765}.

\bibitem[Luo et~al.(2025)Luo, Rechardt, Sun, Nejad, Y{\'a}{\~n}ez, Yilmaz, Lee,
  Cohen, Borghesani, Pashkov, Marinazzo, Nicholas, Salatiello, Sucholutsky,
  Minervini, Razavi, Rocca, Yusifov, Okalova, Gu, Ferianc, Khona, Patil, Lee,
  Mata, Myers, Bizley, Musslick, Bilgin, Niso, Ales, Gaebler, Murty,
  Loued-Khenissi, Behler, Hall, Dafflon, Bao, and Love]{brainbench}
Xiaoliang Luo, Akilles Rechardt, Guangzhi Sun, Kevin~K. Nejad, Felipe
  Y{\'a}{\~n}ez, Bati Yilmaz, Kangjoo Lee, Alexandra~O. Cohen, Valentina
  Borghesani, Anton Pashkov, Daniele Marinazzo, Jonathan Nicholas, Alessandro
  Salatiello, Ilia Sucholutsky, Pasquale Minervini, Sepehr Razavi, Roberta
  Rocca, Elkhan Yusifov, Tereza Okalova, Nianlong Gu, Martin Ferianc, Mikail
  Khona, Kaustubh~R. Patil, Pui-Shee Lee, Rui Mata, Nicholas~E. Myers,
  Jennifer~K Bizley, Sebastian Musslick, Isil~Poyraz Bilgin, Guiomar Niso,
  Justin~M. Ales, Michael Gaebler, N~Apurva~Ratan Murty, Leyla Loued-Khenissi,
  Anna Behler, Chloe~M. Hall, Jessica Dafflon, Sherry~Dongqi Bao, and
  Bradley~C. Love.
\newblock Large language models surpass human experts in predicting
  neuroscience results.
\newblock \emph{Nature Human Behaviour}, 9:\penalty0 305--315, 2025.
\newblock \doi{10.1038/s41562-024-02046-9}.

\bibitem[Majumder et~al.(2024)Majumder, Surana, Agarwal, Mishra, Meena,
  Prakhar, Vora, Khot, Sabharwal, and Clark]{discoverybench}
Bodhisattwa~Prasad Majumder, Harshit Surana, Dhruv Agarwal, Bhavana~Dalvi
  Mishra, Abhijeetsingh Meena, Aryan Prakhar, Tirth Vora, Tushar Khot, Ashish
  Sabharwal, and Peter Clark.
\newblock {DiscoveryBench: Towards Data-Driven Discovery with Large Language
  Models}.
\newblock \emph{arXiv preprint arXiv:2407.01725}, 2024.
\newblock URL \url{https://arxiv.org/abs/2407.01725}.

\bibitem[Mitchener et~al.(2025)Mitchener, Laurent, Andonian, Tenmann,
  Narayanan, Wellawatte, White, Sani, and Rodriques]{bixbench}
Ludovico Mitchener, Jon~M Laurent, Alex Andonian, Benjamin Tenmann, Siddharth
  Narayanan, Geemi~P Wellawatte, Andrew White, Lorenzo Sani, and Samuel~G
  Rodriques.
\newblock {BixBench: a Comprehensive Benchmark for LLM-based Agents in
  Computational Biology}.
\newblock \emph{arXiv preprint arXiv:2503.00096}, 2025.
\newblock URL \url{https://arxiv.org/abs/2503.00096v3}.

\bibitem[Sayers et~al.(2026)Sayers, Bolton, Fine, Kelly, Kim, Landrum, Lathrop,
  Malheiro, Murphy, Phan, Pujar, Trawick, Schneider, and
  Pruitt]{sayers2026ncbi}
Eric~W Sayers, Evan~E Bolton, Anna~M Fine, Christopher Kelly, Sunghwan Kim,
  Melissa Landrum, Stacy Lathrop, Adriana Malheiro, Terence~D Murphy, Lon Phan,
  Shashikant Pujar, Barton~W Trawick, Valerie~A Schneider, and Kim~D Pruitt.
\newblock Database resources of the {National Center for Biotechnology
  Information} in 2026.
\newblock \emph{Nucleic Acids Research}, 54\penalty0 (D1):\penalty0 D20--D27,
  2026.
\newblock \doi{10.1093/nar/gkaf1060}.

\bibitem[Siegel et~al.(2024)Siegel, Kapoor, Nadgir, Stroebl, and
  Narayanan]{corebench}
Zachary~S. Siegel, Sayash Kapoor, Nitya Nadgir, Benedikt Stroebl, and Arvind
  Narayanan.
\newblock {CORE-Bench: Fostering the Credibility of Published Research Through
  a Computational Reproducibility Agent Benchmark}.
\newblock \emph{arXiv preprint arXiv:2409.11363}, 2024.
\newblock URL \url{https://arxiv.org/abs/2409.11363}.

\bibitem[Silberzahn et~al.(2018)Silberzahn, Uhlmann, Martin, Anselmi, Aust,
  Awtrey, Bahn{\'i}k, Bai, Bannard, Bonnier, Carlsson, Cheung, Christensen,
  Clay, Craig, Dalla~Rosa, Dam, Evans, Flores~Cervantes, Fong, Gamez-Djokic,
  Glenz, Gordon-McKeon, Heaton, Hederos, Heene, Hofelich~Mohr, H{\"o}gden, Hui,
  Johannesson, Kalodimos, Kaszubowski, Kennedy, Lei, Lindsay, Liverani, Madan,
  Molden, Molleman, Morey, Mulder, Nijstad, Pope, Pope, Prenoveau, Rink,
  Robusto, Roderique, Sandberg, Schl{\"u}ter, Sch{\"o}nbrodt, Sherman, Sommer,
  Sotak, Spain, Sp{\"o}rlein, Stafford, Stefanutti, Tauber, Ullrich, Vianello,
  Wagenmakers, Witkowiak, Yoon, and Nosek]{Silberzahn2018ManyAO}
R.~Silberzahn, E.~L. Uhlmann, D.~P. Martin, P.~Anselmi, F.~Aust, E.~Awtrey,
  {\v{S}}.~Bahn{\'i}k, F.~Bai, C.~Bannard, E.~Bonnier, R.~Carlsson, F.~Cheung,
  G.~Christensen, R.~Clay, M.~A. Craig, A.~Dalla~Rosa, L.~Dam, M.~H. Evans,
  I.~Flores~Cervantes, N.~Fong, M.~Gamez-Djokic, A.~Glenz, S.~Gordon-McKeon,
  T.~J. Heaton, K.~Hederos, M.~Heene, A.~J. Hofelich~Mohr, F.~H{\"o}gden,
  K.~Hui, M.~Johannesson, J.~Kalodimos, E.~Kaszubowski, D.~M. Kennedy, R.~Lei,
  T.~A. Lindsay, S.~Liverani, C.~R. Madan, D.~Molden, E.~Molleman, R.~D. Morey,
  L.~B. Mulder, B.~R. Nijstad, N.~G. Pope, B.~Pope, J.~M. Prenoveau, F.~Rink,
  E.~Robusto, H.~Roderique, A.~Sandberg, E.~Schl{\"u}ter, F.~D. Sch{\"o}nbrodt,
  M.~F. Sherman, S.~A. Sommer, K.~Sotak, S.~Spain, C.~Sp{\"o}rlein,
  T.~Stafford, L.~Stefanutti, S.~Tauber, J.~Ullrich, M.~Vianello, E.-J.
  Wagenmakers, M.~Witkowiak, S.~Yoon, and B.~A. Nosek.
\newblock Many analysts, one data set: Making transparent how variations in
  analytic choices affect results.
\newblock \emph{Advances in Methods and Practices in Psychological Science},
  1\penalty0 (3):\penalty0 337--356, 2018.
\newblock \doi{10.1177/2515245917747646}.

\bibitem[Starace et~al.(2025)Starace, Jaffe, Sherburn, Aung, Chan, Maksin,
  Dias, Mays, Kinsella, Thompson, Heidecke, Glaese, and Patwardhan]{paperbench}
Giulio Starace, Oliver Jaffe, Dane Sherburn, James Aung, Jun~Shern Chan, Leon
  Maksin, Rachel Dias, Evan Mays, Benjamin Kinsella, Wyatt Thompson, Johannes
  Heidecke, Amelia Glaese, and Tejal Patwardhan.
\newblock {PaperBench: Evaluating AI's Ability to Replicate AI Research}.
\newblock \emph{arXiv preprint arXiv:2504.01848}, 2025.
\newblock URL \url{https://arxiv.org/abs/2504.01848}.

\bibitem[Tan and Tian(2026)]{medresearchbench}
Shuping Tan and Zhanxiao Tian.
\newblock {MedResearchBench: A Multi-Domain Benchmark for Evaluating AI
  Research Agents on Clinical Medical Research}.
\newblock \emph{medRxiv}, 2026.
\newblock \doi{10.64898/2026.03.30.26349749}.
\newblock URL \url{https://doi.org/10.64898/2026.03.30.26349749}.

\bibitem[Wang et~al.(2026)Wang, Bai, Luo, Su, Sun, Yu, Liu, Zhou, Cardie,
  Dredze, Hu, and Xing]{firebench}
Zhen Wang, Fan Bai, Zhongyan Luo, Jinyan Su, Kaiser Sun, Xinle Yu, Jieyuan Liu,
  Kun Zhou, Claire Cardie, Mark Dredze, Zhiting Hu, and Eric~P. Xing.
\newblock {FIRE-Bench}: Evaluating {AI} agents on the rediscovery of scientific
  insights.
\newblock \emph{arXiv preprint arXiv:2602.02905}, 2026.
\newblock URL \url{https://arxiv.org/abs/2602.02905}.

\bibitem[Wang et~al.(2025)Wang, Danek, and Sun]{biodsa}
Zifeng Wang, Benjamin Danek, and Jimeng Sun.
\newblock {BioDSA-1K: Benchmarking Data Science Agents for Biomedical
  Research}.
\newblock \emph{arXiv preprint arXiv:2505.16100}, 2025.
\newblock URL \url{https://arxiv.org/abs/2505.16100}.

\bibitem[White et~al.(2024)White, Dooley, Roberts, Pal, Feuer, Jain,
  Shwartz-Ziv, Jain, Saifullah, Dey, Shubh-Agrawal, Sandha, Naidu, Hegde,
  LeCun, Goldstein, Neiswanger, and Goldblum]{livebench}
Colin White, Samuel Dooley, Manley Roberts, Arka Pal, Ben Feuer, Siddhartha
  Jain, Ravid Shwartz-Ziv, Neel Jain, Khalid Saifullah, Sreemanti Dey,
  Shubh-Agrawal, Sandeep~Singh Sandha, Siddartha Naidu, Chinmay Hegde, Yann
  LeCun, Tom Goldstein, Willie Neiswanger, and Micah Goldblum.
\newblock {LiveBench: A Challenging, Contamination-Limited LLM Benchmark}.
\newblock \emph{arXiv preprint arXiv:2406.19314}, 2024.
\newblock URL \url{https://arxiv.org/abs/2406.19314}.
\newblock ICLR 2025.

\bibitem[Wu et~al.(2024)Wu, Pan, Xie, Zhou, Zhao, Ma, Du, Mao, Luu, and
  Wang]{antileak}
Xiaobao Wu, Liangming Pan, Yuxi Xie, Ruiwen Zhou, Shuai Zhao, Yubo Ma, Mingzhe
  Du, Rui Mao, Anh~Tuan Luu, and William~Yang Wang.
\newblock {AntiLeakBench: Preventing Data Contamination by Automatically
  Constructing Benchmarks with Updated Real-World Knowledge}.
\newblock \emph{arXiv preprint arXiv:2412.13670}, 2024.
\newblock URL \url{https://arxiv.org/abs/2412.13670}.

\bibitem[Xu et~al.(2024)Xu, Wang, Fan, and Liu]{xu2024benchmarking}
Ruijie Xu, Zengzhi Wang, Run-Ze Fan, and Pengfei Liu.
\newblock Benchmarking benchmark leakage in large language models.
\newblock \emph{arXiv preprint arXiv:2404.18824}, 2024.
\newblock URL \url{https://arxiv.org/abs/2404.18824}.

\bibitem[Ye et~al.(2026)Ye, Chen, Tu, Liu, Xiong, Schmidgall, and
  Bitterman]{proofoftime}
Bingyang Ye, Shan Chen, Jingxuan Tu, Chen Liu, Zidi Xiong, Samuel Schmidgall,
  and Danielle~S. Bitterman.
\newblock Proof of time: A benchmark for evaluating scientific idea judgments.
\newblock \emph{arXiv preprint arXiv:2601.07606}, 2026.
\newblock URL \url{https://arxiv.org/abs/2601.07606}.

\bibitem[Ye et~al.(2025)Ye, Yuan, Cooray, Dillmann, Roque, Baron, Frank,
  Martin-Alvarez, Koblischke, Qu, Yang, Wechsler, and Ciuca]{replicationbench}
Christine Ye, Sihan Yuan, Suchetha Cooray, Steven Dillmann, Ian L.~V. Roque,
  Dalya Baron, Philipp Frank, Sergio Martin-Alvarez, Nolan Koblischke, Frank~J
  Qu, Diyi Yang, Risa Wechsler, and Ioana Ciuca.
\newblock {ReplicationBench: Can AI Agents Replicate Astrophysics Research
  Papers?}
\newblock \emph{arXiv preprint arXiv:2510.24591}, 2025.
\newblock URL \url{https://arxiv.org/abs/2510.24591}.

\end{thebibliography}

\clearpage
\appendix

\section{Leakage controls}\label{app:controls}

\subsection{Pre-training and post-training data}
Leaked test data distorts benchmark comparisons \citep{xu2024benchmarking}. Every task comes from a paper first made public after June 2026, the latest developer-reported knowledge cutoff among the evaluated models (Table~\ref{tab:cutoffs}).

\begin{table}[h]
\centering\small
\caption{\textbf{Evaluated models} and their knowledge cutoffs as published by each developer.}
\label{tab:cutoffs}
\begin{tabular}{@{}lllp{5.2cm}@{}}
\toprule
Model & Developer & Weights & Knowledge cutoff \\
\midrule
Claude Opus 5.5 & Anthropic & closed & June 2026 \\
GPT-6 Astra & OpenAI & closed & 30 April 2026 \\
GPT-6 Luna & OpenAI & closed & 18 May 2026 \\
Gemini 3.8 Flash & Google & closed & March 2026 \\
GLM 5.3 & Z.ai & open & January 2026 \\
MiMo V2.6 Pro & Xiaomi & open & November 2025 \\
DeepSeek V4.1 Flash & DeepSeek & open & December 2025 \\
Qwen3.8 27B & Alibaba (Qwen) & open & September 2025  \\
\bottomrule
\end{tabular}
\end{table}
 We date each paper by its earliest appearance across preprint servers, journal online-first dates and print issues, which excluded 95 candidates with earlier preprints. We use real post-cutoff papers rather than instructing a model to ignore what it knows, because prompted cutoffs fail, since models still use later knowledge when it is causally relevant \citep{rewindtime}, and instructions to respect historical boundaries do not prevent recall of memorised values \citep{memorizationecon}.

\subsection{In-environment leakage controls}
Agents need the literature for variable definitions, codebooks and the conventions of earlier studies, so we restrict search by date rather than remove it, since unrestricted retrieval leaks answers \citep{sciconbench,anthropic2026evalawareness}. The tools apply the cutoff in each upstream query, re-check every returned date and remove forward links such as citing papers. We utilize a re-check  because we found that upstream dates are unreliable (i.e., given a date of January 1st 2026 when there is missing information about the exact date and month). Each task also blocklists the seed paper's DOI, PubMed and PubMed Central identifiers and publisher URLs, following PaperBench's use of per-paper lists of prohibited repositories \citep{paperbench}. Web search is disabled, the shell reaches only allowlisted data and package hosts, and each run's sandbox contains only its own files, with the paper's findings stored outside it. Across \leakToolCalls{} tool calls, \leakRecordsAfterCutoff{} of the \leakRecordsReturned{} records the tools returned were dated after the cutoff.

\subsection{Post-processing}
After each run, an audit re-resolves every identifier the agent encountered, including those it cited from memory, against Europe PMC and Crossref, and flags any first made public after the cutoff.

\section{Failure cases}\label{app:fail}
\begin{figure}[h]
\centering
\includegraphics[width=\linewidth]{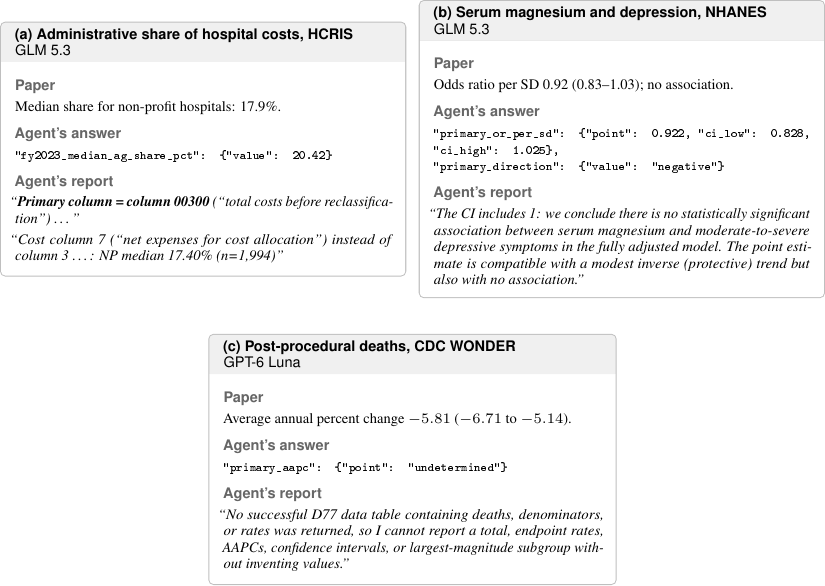}
\caption{\textbf{Failure cases}, quoted from the agents' answers and reports. (a) The agent reports the pre-adjustment cost share as its answer and the value matching the paper only as a sensitivity analysis. (b) The report concludes no association, but the structured answer records the direction of the point estimate. (c) The agent obtains no data and declines to estimate.}
\label{fig:fail}
\end{figure}

Figure~\ref{fig:fail} shows detailed qualitative examples of failure cases, each quoting the agent's structured answer and report alongside the paper's result.

\section{Predictability of post-cutoff findings}\label{app:predict}
Post-cutoff findings are largely predictable from the research question alone, and the prediction does not depend on the model recognising the paper. We gave GPT-6 Sol only the neutral question for each of the \blindN{} shortlisted papers, all first made public between July and September 2026, after its cutoff. Without tools, it returned a direction, a significance call, a numeric estimate, a confidence and whether it knew the finding, and we compared these with the primary claim parsed from each abstract (Table~\ref{tab:predict}).

\begin{table}[h]
\centering\small
\caption{\textbf{Blind guesses on the \blindN{} shortlisted papers.} Rates are over the papers for which each quantity is defined (count in grey); -- marks cells with fewer than 15 papers.}
\label{tab:predict}
\resizebox{\linewidth}{!}{\begin{tabular}{@{}lrcccc@{}}
\toprule
 & & \multicolumn{2}{c}{Direction and significance matched} & Significant effect & Numeric guess \\
\cmidrule(lr){3-4}
Papers & $n$ & non-null findings & null findings & predicted for nulls & inside paper CI \\
\midrule
All shortlisted papers & 2{,}131 & 83\% \textcolor{gray}{(1673)} & 16\% \textcolor{gray}{(101)} & 61\% \textcolor{gray}{(104)} & 52\% \textcolor{gray}{(1313)} \\
\midrule
Model reports it does not know the finding & 1{,}814 & 81\% \textcolor{gray}{(1399)} & 17\% \textcolor{gray}{(90)} & 57\% \textcolor{gray}{(93)} & 51\% \textcolor{gray}{(1090)} \\
Model is unsure whether it knows it & 317 & 95\% \textcolor{gray}{(274)} & -- & -- & 54\% \textcolor{gray}{(223)} \\
\midrule
First public July 2026 & 913 & 85\% \textcolor{gray}{(718)} & 12\% \textcolor{gray}{(41)} & 69\% \textcolor{gray}{(42)} & 50\% \textcolor{gray}{(578)} \\
First public August 2026 & 698 & 82\% \textcolor{gray}{(555)} & 11\% \textcolor{gray}{(28)} & 63\% \textcolor{gray}{(30)} & 54\% \textcolor{gray}{(419)} \\
First public September 2026 & 520 & 83\% \textcolor{gray}{(400)} & 25\% \textcolor{gray}{(32)} & 47\% \textcolor{gray}{(32)} & 51\% \textcolor{gray}{(316)} \\
\midrule
Data: NHANES & 462 & 87\% \textcolor{gray}{(418)} & 12\% \textcolor{gray}{(16)} & 56\% \textcolor{gray}{(16)} & 59\% \textcolor{gray}{(374)} \\
Data: CHARLS & 201 & 93\% \textcolor{gray}{(192)} & -- & -- & 55\% \textcolor{gray}{(159)} \\
Data: MIMIC & 186 & 90\% \textcolor{gray}{(167)} & -- & -- & 48\% \textcolor{gray}{(155)} \\
Data: SEER & 93 & 83\% \textcolor{gray}{(75)} & -- & -- & 60\% \textcolor{gray}{(67)} \\
Data: GWAS / MR & 67 & 62\% \textcolor{gray}{(58)} & -- & -- & 36\% \textcolor{gray}{(56)} \\
Data: FAERS & 37 & 97\% \textcolor{gray}{(29)} & -- & -- & 30\% \textcolor{gray}{(27)} \\
Data: all other sources & 1{,}085 & 79\% \textcolor{gray}{(734)} & 20\% \textcolor{gray}{(51)} & 54\% \textcolor{gray}{(52)} & 48\% \textcolor{gray}{(475)} \\
\bottomrule
\end{tabular}
}
\end{table}

The model matched the direction and significance of \predKnownNoRate\% of non-null findings it reported not knowing, and of \predUnsureRate\% of those it was unsure about. The rate is flat across the three months of first publication (\predMonthMin--\predMonthMax\%). Predictability is highest for heavily mined sources such as NHANES, CHARLS, MIMIC and FAERS and lowest for Mendelian randomisation, whose instruments and outcome GWAS vary between studies. The guesses also reproduce publication bias: the model predicted a significant effect for \blindSigNullPct\% of papers whose finding was null (\blindSigNullK{} of \blindSigNullN) and matched only \blindNullPct\% of null findings.

\section{Claim-level outcomes}\label{app:claims}
\begin{figure}[h]
\centering
\includegraphics[width=\linewidth]{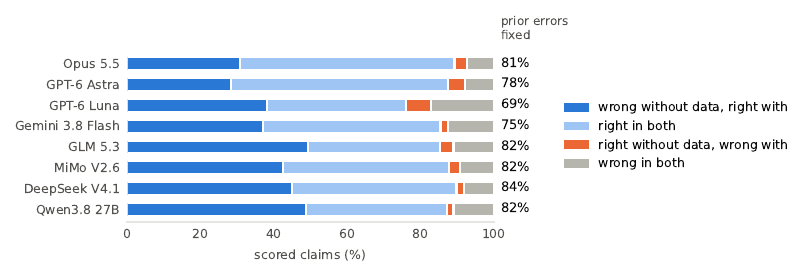}
\caption{\textbf{What the data changes}: outcome of every scored claim from the answer without data to the answer with data, per model; the number on the right is the share of prior errors the data corrected.}
\label{fig:pp}
\end{figure}

Shown in Figure~\ref{fig:pp}, access to data corrects between \ppFixedMin\% and \ppFixedMax\% of each model's prior errors, while minimally reversing the claims its prior got right.

\section{Screening prompts}\label{app:prompts}
\label{app:blind}
\begin{figure}[h]
\centering
\includegraphics[width=0.85\linewidth]{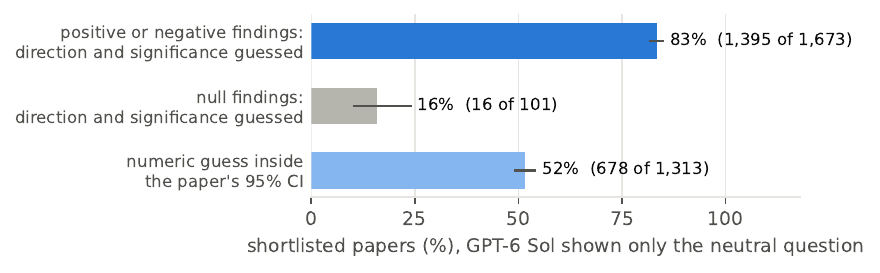}
\caption{\textbf{Recall screen.} Share of shortlisted papers whose direction and significance GPT-6 Sol matched from the question alone, by the paper's finding, and share of numeric guesses inside the paper's 95\% CI, with 95\% Wilson intervals.}
\label{fig:blind}
\end{figure}

We report the LLM prompts used to filter publications at scale for tasks fit for \bench{}. Stages 1 and 2 ran with GPT-6 Luna over every candidate record. The subsequent stages used GPT-6 Sol, over the \blindN{} shortlisted papers. As shown in Figure~\ref{fig:blind}, the blind guess sees only the neutral question and powers the recall screen, predicting a significant effect for \blindSigAllPct\% of papers and for \blindSigNullPct\% of those with null findings (\blindSigNullK{} of \blindSigNullN), so findings it recovered were demoted.

\subsection{Stage 1: study type}
\begin{lstlisting}
Classify the PubMed record as JSON. Did the authors generate ANY new primary data, or only analyze data that already existed? Abstract outranks accessions and publication types. Apply steps IN ORDER; the first match sets `primary_data_collection` (pdc) and `design`.

NEW cues (authors did it): recruited, enrolled (unless "retrospective" also appears), randomized, placebo, informed consent, healthy controls, volunteers, "participants", questionnaires/interviews the authors administered, specimens collected, "we measured/sequenced/imaged", mice/cells treated, synthesized, deployed (cameras, sensors). "Prospective" is NEW only if the authors themselves enrolled participants.

EXISTING cues: retrospective(ly) anywhere in title, MeSH ("Retrospective Studies") or abstract (decisive: always step 10, never unclear), medical records, database, registry, claims, EHR, biobank, "data were obtained/downloaded from", publicly available, target trial emulation, population-based, GWAS summary statistics, bioinformatics, systematic search, bibliometric; any named cohort, survey or public dataset in ANY field, including its assays (EXISTING unless participants were recruited for THIS paper); vendor/device data recorded before the study; data the authors collected earlier under another protocol. Measurements, contacts or questionnaires made for THIS paper are NEW.

# Decision steps

1. No results or no data analysis: future-tense protocol ("will be enrolled"), editorial, guideline, consensus statement (including Delphi-based), classification or framework proposal with no cohort and no N, bench-protocol chapter (Methods in Molecular Biology, Bio-protocol, "here we describe the method") that lists procedures but reports no results: other_non_research, pdc=none. Protocol verbs are not evidence of new data; look for an N, a measurement or a result. Future tense overrides PUBLICATION TYPES and trial accessions. Data descriptors are step 4 or 7.

2. Descriptive report of <=10 patients the authors treated, or titled "case series", no comparison or effect estimate: case_report_or_series, pdc=new_human_data. Retrospective comparisons of any N are step 10.

3. The headline is the performance of a tool, kit, assay, device, algorithm, chatbot or rater (coverage, duplication rate, LOD, recovery, accuracy or AUC versus a reference, ICC/kappa/Bland-Altman agreement, "our approach outperforms" on benchmark data), or the subject matter is non-biomedical (finance, markets, traffic, industrial processes): methods_or_software regardless of any new samples or participants used as test material; pdc=new_* if that test material was newly acquired, else none. This step outranks 4 and 7.

4. Authors generated measurements on animals (including pets, livestock), plants, cells, chemicals, materials or environments (chemistry, field work) with functional experiments or no pre-existing dataset analyzed: lab_or_animal_experiment, pdc=new_lab_or_animal_data, even without a collection verb. Human group comparisons whose measurements are bench assays on specimens obtained for this paper (biopsy IHC, serum panels, histology) are lab_or_animal_experiment with pdc=new_human_data. A deposited GEO/SRA accession is NEW; "downloaded" accessions are existing sources.

5. Existing-dataset analysis whose ONLY new data is validation (qPCR, western blot, IHC on a few samples, no functional assay) or a small new patient cohort confirming the finding: secondary_analysis_existing_data, pdc=new_lab_or_animal_data (new_human_data for the cohort), mixed_minor_validation=true.

6. Authors assigned an intervention or experimental condition to humans and report their own results: randomized, placebo, task conditions, crossover, repeated measures before/after an author-applied manipulation (anaesthetic block, load, sham, fasting, posture), n-of-1: rct_or_interventional, pdc=new_human_data. Outranks step 7 whenever the authors controlled the exposure. Prespecified substudies reporting protocol measurements are trial reports; analyses labelled secondary, post hoc or exploratory are step 10 (trial_data_reuse).

7. Authors recruited or followed participants for this paper, measured them, used research-only instruments (psychometric scales, neuropsychological batteries, patient-reported scales: records cannot hold these, never unclear), ran their own surveys/interviews/Delphi, or assayed stored cohort specimens: prospective_primary_cohort (qualitative_or_survey_primary for surveys/interviews); pdc=new_human_data. Non-patient samples (workers, students) and cross-sectional studies of "participants" belong here even without a recruitment verb.

8. Literature synthesis (systematic search, bibliometric): with a pooled estimate (pooled OR/HR/SMD, I2, network or IPD meta-analysis) meta_analysis_or_pooled (is_secondary_data_study=true), else systematic_or_narrative_review; pdc=none.

9. Model fed only by published parameters (Markov, cost-effectiveness), or simulation of method performance even if resampling public data: modeling_or_simulation. New algorithm, software, assay or scale benchmarked on public or test-bed data ("we propose [tool]"): methods_or_software; pdc=none (never unclear) unless the authors acquired new test data (then new_*, same design). Tie-break: a model fitted to a record-level clinical, biological or population-health dataset (risk score, hub genes, clinical prediction model with AUC headline) is step 10; only a new algorithm compared against other algorithms, or a non-biomedical subject, stays methods_or_software.

10. Analysis of existing data only (EXISTING cue, no NEW cue): secondary_analysis_existing_data, pdc=none. Also "consecutive patients treated between [past years]", routine clinical/imaging/lab data with analysis verbs, past date ranges and no consent/specimen/intervention language, even if it says "enrolled" or "cohort study" (confidence 0.7-0.85).

11. Otherwise: pdc=unclear, design=secondary_analysis_existing_data, confidence<=0.5, one descriptive data_sources entry with access=unknown. Wet-lab, animal, plant or material work never reaches this step.

# Field rules

- is_secondary_data_study: true ONLY when pdc=none AND design is secondary_analysis_existing_data or meta_analysis_or_pooled; otherwise false.
- mixed_minor_validation: true iff design=secondary_analysis_existing_data AND pdc is new_* (step 5).
- data_sources: pre-existing datasets only; never the authors' own participants, samples or animals; empty for steps 1, 2, 4, 6, 7 and parameter-only models. REQUIRED (>=1 entry) when design is secondary_analysis_existing_data or meta_analysis_or_pooled; if the abstract does not name the dataset, one descriptive entry with access=unknown. Primary-result source FIRST (name it in the rationale), validation-only sources after. Type and access:
  open: NHANES, BRFSS, CDC WONDER, GBD [public_dataset_or_survey]; ClinicalTrials.gov [registry]; GEO, SRA, TCGA/GDC open tier, GTEx, all GWAS summary statistics (GWAS Catalog, IEU OpenGWAS) [omics_repository]; FAERS/openFDA [pharmacovigilance]; published literature [literature].
  registration: CHARLS, HRS, ELSA, SHARE, CLSA, DHS/MICS [public_dataset_or_survey]; SEER [registry]; MIMIC, eICU [ehr_or_claims]; BioLINCC, YODA, Vivli trial IPD [trial_data_reuse].
  restricted: individual-level UK Biobank, All of Us, dbGaP-controlled [biobank]; Medicare, TriNetX, national claims [ehr_or_claims]; national or government registers of any country (Nordic, Norwegian, Danish, cancer, cause-of-death, vital statistics, device registers): [registry], never unknown, never institutional_records; VigiBase, JADER [pharmacovigilance]; Web of Science, Scopus [literature]; trial IPD unless BioLINCC/YODA/Vivli named [trial_data_reuse]; named cohorts, consortium data, vendor/device data [other]; hospital, clinic or single-center chart data ONLY [institutional_records], never unknown.
  unknown: otherwise. A source on the lists above always takes its listed type and tier.
- focus: FIRST, if design is not one of secondary_analysis_existing_data, meta_analysis_or_pooled, prospective_primary_cohort, modeling_or_simulation, focus MUST be other, no exceptions even when ORs or betas are reported. Otherwise: genetic_gwas_or_mr; omics_bioinformatics (DEGs, network pharmacology); pharmacovigilance; health_services_or_cost; diagnostic_accuracy (test vs reference standard); prediction_model (scored model whose AUC/C-index is the headline); treatment_effect_observational (treatments/policies compared in non-randomised data or trial re-analysis, e.g. target trial emulation); descriptive_epidemiology (prevalence, trends, no exposure-outcome estimate); association_or_etiology (any other exposure-outcome estimate, including "predictors" reported as ORs/HRs).
- confidence: 0.9-1.0 explicit quoted cue; 0.7-0.85 inferred; <=0.5 unclear.
- rationale: <=30 words quoting the deciding cue and the primary source.

# Worked examples (cue -> pdc / design / focus / is_secondary; sources)

- "Multicentre cohort, N=2,148 treated 2015-2022; target trial emulation" -> none / secondary_analysis_existing_data / treatment_effect_observational / true; [multicentre records, institutional_records, restricted]; confidence 0.8, NOT unclear
- "GSE18781 downloaded from GEO; hub genes validated by qPCR in patient samples" -> new_lab_or_animal_data / secondary_analysis_existing_data / omics_bioinformatics / false; mixed=true; [GSE18781, omics_repository, open]
- "Three WES capture kits compared on 8 DNA samples; coverage uniformity" -> new_lab_or_animal_data / methods_or_software / other / false; []
\end{lstlisting}
\subsection{Stage 2: scorable claim and obtainable data}
\begin{lstlisting}
You screen PubMed records for an environment where an agent, given a public dataset at a fixed version and a neutral question, must re-derive the paper's headline result. Judge from the record alone; fill every field. Score only in Step 7.

## 1. Gate: analysis of pre-existing data?
Fails if PUBLICATION TYPES or TITLE has Review, Meta-Analysis of published studies (not pooled GWAS summary statistics or GEO series), Editorial/Letter/Comment, Case Report, Protocol; if the authors generated data (recruited/enrolled/randomized participants, questionnaires, animal/cell/in-vitro work, assays, imaging or sequencing for this study; bioinformatics screens whose headline is wet-lab validation; dosimetric or treatment-planning studies where the authors generated new plans or simulations); if the headline is the performance of an AI model, LLM, software tool or device evaluated on the data (a tool evaluation, not a data analysis); or no empirical dataset. DATABANK ACCESSIONS: NCT/DRKS/ChiCTR = the authors' own trial; GSE/PRJNA/figshare/Dryad usually = data they DEPOSITED unless the abstract says it was downloaded.
(A) Gate FAILS: use ALL defaults: booleans false, data_publicly_obtainable "no", type "other", direction "na", strings "" (except concerns), confidence >= 0.9.
(B) Gate PASSES but data is "no" (retrospective review of hospital, registry, EHR, imaging or genetic-test records): fill primary_claim, data_sources and data_version_stated normally; single_primary_question follows Step 5; score 1.
Bibliometric/publication-count studies: gate passes (PubMed is public) but single_primary_question=false (multi-aim descriptive), score 2.

## 2. Headline claim -> primary_claim
First rule that applies: (a) the outcome labelled "primary" (several: the first listed); (b) the exposure-outcome pair in the TITLE (first-named exposure; bidirectional MR: forward direction); (c) the first exposure-outcome result in RESULTS, skipping sample descriptors (N, mean age, % with a trait). ONE exposure only, the first named: for "predictors of / risk factors for X" papers take the first predictor and its own sign; never concatenate predictors. Never a subgroup, secondary, sensitivity or crude result, and never by significance. "Prevalence and risk factors of X": the prevalence. Shape plus estimate: the estimate is the claim; shape goes in concerns. If text and numbers disagree, follow the numbers. When the headline comes from rule (c), confidence <= 0.6 (hard cap).
type: prevalence/rate/APC -> estimate; any AUC, including a pre-existing score (SOFA) validated in public data -> auc. value_text: numbers verbatim with CI/p, adjusted over crude; set: selection criterion plus members or their count; "" if none. direction = sign of the association: positive (OR/HR/RR>1), negative (HR<1, even for treatment vs control), null (CI includes null), mixed (ONE exposure whose sign reverses by subgroup or timepoint, never several predictors with different signs), na (auc, set, rank, count, categorical, other).

## 3. answer_objective, null_or_negative_result
answer_objective=true only if value_text has a number or an enumerated set ("significantly higher" alone is not). Borderline = 0.01<=p<0.05 or CI bound within 0.05 of null; write "borderline" in robustness_signals.
null_or_negative_result=true ONLY if direction is null, or the conclusion states the exposure did NOT have the expected or hypothesised effect. A significant protective or beneficial effect (HR/OR<1) is direction=negative with null_or_negative_result=false. Never lowers the score.

## 4. Data
data_sources: names as written with years/version. Grade the WORST source the primary claim needs (public discovery + private validation: grade discovery; public data merged with data the authors compiled: "no").
yes: NHANES, NHIS, BRFSS, MEPS, CDC WONDER, GBD, FAERS/openFDA, GEO, TCGA/GDC open, GTEx, CGGA, cBioPortal, GWAS Catalog, IEU OpenGWAS, any published GWAS summary statistics, ClinicalTrials.gov records, PubMed/OpenAlex metadata. MR: grade the summary statistics, not the cohort (UK Biobank summary statistics = yes).
registration: SEER (always "registration", never "no"), MIMIC, eICU, HRS, CHARLS, ELSA, SHARE, BioLINCC.
no: UK Biobank, All of Us, dbGaP controlled, national registers/insurance (Nordic, Korea, Taiwan, CPRD, JAGES), US claims/registries (Medicare/CMS, HCUP NIS/NRD/KID, NCDB, NSQIP, NTDB), commercial (Optum, TriNetX), consortium or multicentre cohort data (CENTER-TBI, IBBC, PIERS, EXCEL), QI or registry initiatives (MAQI2), national HMIS/DHIS2 extracts, research-group cohorts, trial data without a named repository (an NCT number is not one), "available on reasonable request", ANY hospital/medical-records database, any UNNAMED source ("a nationwide database", city-level data the authors compiled), except unnamed GWAS summary statistics.
unknown: a named repository or portal that LOOKS public but is not on the yes list (Rembrandt/GlioVis), or GWAS summary statistics from unnamed studies.
data_version_stated=true only if a cycle, year range, release or accession is given ("MIMIC-IV" alone = false).

## 5. single_primary_question, question_neutral
false (forces score 2) if ANY: many exposures or outcomes with no labelled primary; multi-aim descriptive paper ("burden of X", incidence AND survival AND disparities, "characterize reporting patterns") - never carve out one rate as the primary; >=3 independent analytic arms (MR + colocalisation + pharmacovigilance); the deliverable is a model, risk score, prognostic signature, hub-gene/PPI pipeline, database/web platform, software or method. Bidirectional MR (forward direction), "prevalence and risk factors" designs (primary = the prevalence), secondary outcomes, subgroups and sensitivity analyses do NOT break singleness. Omics set claims: a screen of a named gene family or a DEG list on TCGA/GEO/CGGA whose result is an enumerated set of genes with a stated criterion (direction of survival association, FDR<0.05) is ONE question: type=set, value_text=criterion plus members, answer_objective=true; the hub-gene/PPI exclusion applies only when the deliverable is a network, signature score or wet-lab validation.
question_neutral: "" when single_primary_question is false or the score is 1; never write "Not applicable". Otherwise sentence 1 = population, dataset + version, exposure, outcome, main adjustment; sentence 2 = "Report ..." with exactly ONE estimand and its format: "Report the adjusted OR with 95% CI and state whether an association is present". Never "for each outcome" or "each measure" (that means single_primary_question is false). Never include result numbers, direction words (increase, reduce, protective, risk factor), the significant subgroup or feature, or the title.

## 6. analysis_complexity
Rate ONLY the primary claim's analysis; ignore auxiliary ML/SHAP, splines, nomograms, mechanistic add-ons and a source's own modelling. low: regression, survival, propensity/IPTW, joinpoint/APC, disproportionality, two-sample IVW MR, differential expression. medium: DiD/target-trial emulation, data-derived thresholds, MR with colocalisation, ML with cross-validation. high: raw read alignment/variant calling, GWAS from genotypes, deep learning on images/waveforms, proprietary software.

## 7. environment_buildability: the FIRST line that applies
1: gate failed, or data "no", or answer_objective false.
2: single_primary_question false (even with open data), or complexity high.
3: data "unknown"; or estimate with neither CI nor p; or set with no stated criterion; or ABSTRACT empty or truncated mid-sentence (then confidence<=0.3).
4: data "registration"; or complexity medium; or borderline; or data_version_stated false; or N<1000 and not nationally representative.
5: otherwise.
Nothing else changes it. If answer_objective is false or data is "no" the score is 1 regardless of anything below. verdict is a pure function of the score: 5-4 strong, 3 possible, 2 weak, 1 no.

## 8. Remaining fields
robustness_signals (<=30 words): N, p/CI, sensitivity analyses, replication, "borderline"; "none stated" if absent. concerns (<=40 words): never empty when verdict != strong; start with the step that failed ("Step 1: authors enrolled participants", "Step 4: single-centre EHR, not public"). Cite only rules written in this document; if a study type is not covered, say so and apply the best-fit step. confidence: >=0.9 only when every step had an explicit cue; NEVER above 0.6 when data is "unknown" or the headline came from rule (c).

## Examples
"NHANES 2009-2014, BRI vs periodontitis, OR 1.053 (1.016-1.092), N=10,790" -> yes, estimate, positive, low, 5 strong. "MIMIC-IV, PIV vs 60-day mortality, HR 1.20 (1.01-1.44) above a U-shape inflection" -> registration, estimate, positive, borderline, 4 strong. "SEER Texas endometrial cancer incidence by age, race, histology plus survival" -> single_primary_question=false, 2 weak. "Bidirectional MR on unnamed GWAS summary statistics, forward OR 0.99 (0.97-1.01)" -> unknown, null, null_or_negative_result=true, 3 possible.
\end{lstlisting}
\subsection{Recall screen: blind guess from the question alone}
\begin{lstlisting}
You are a senior clinical epidemiologist with excellent recall of the biomedical literature up to your knowledge cutoff. You will be shown ONLY a research question (and the exposure, outcome and population it concerns), with no data and no paper. Answer as if a colleague asked you to bet on the result before the analysis is run.

Give: the direction you expect (positive = higher exposure goes with higher outcome/risk; negative = lower; null = no association; na for counts, prevalences, sets or AUCs where direction does not apply), whether you expect it to be statistically significant at 0.05 in a study of typical size for that dataset, your single best numeric guess in the units the question implies, your probability that direction AND significance are right, and whether you recognise this specific finding from literature you have seen (yes / unsure / no). For prevalences, rates, AUCs and counts, the numeric guess is the whole answer: be as precise as you can.

Do not hedge into "null" merely because you are unsure; give the answer you would actually bet on. Base it on what you know about the exposure-outcome pair, the population, and the typical sample size and analysis for the named dataset.
\end{lstlisting}
\subsection{Review classifier}
\begin{lstlisting}
You are the lead designer of a benchmark of controlled data-analysis environments. In each environment an AI agent receives (a) the exact public dataset file(s) a paper used, at a fixed version, and (b) a neutral research question that does not reveal the result. The agent must analyze the data and report the headline estimate. It is scored on reproducing the paper's DIRECTION and SIGNIFICANCE (and, more loosely, whether its estimate overlaps the paper's confidence interval). A separate "no-data" run checks whether a model can state the answer from memory; items it passes are discarded as leaked.

You read a PubMed record plus fallible screener output, a deterministic arithmetic check of the reported estimate, and a BLIND GUESS made by a strong model that saw only the neutral question. Decide whether this paper would make a good controlled environment. Think like an experimentalist: the environment is only useful if the answer is determined by the data plus a conventional analysis, is not guessable from prior knowledge, and can be scored against a precise number. Judge the estimand, not the abstract's writing style. Distinguish two kinds of things you cannot see in an abstract:
- ROUTINE checks that every environment build performs anyway and that do NOT lower the verdict: the exact cycle/submission/release of a source that is versioned by nature (NHANES cycles, SEER submission year, MIMIC-IV version, FinnGen release, GBD round) once the source is named; the full covariate list; the CI when direction and p are given; the cause-of-death or code-selection rule; software versions; weighting details. List them in needs_full_text and move on.
- DECISIVE unknowns that could flip the verdict and therefore cap it at maybe: whether the data tier is actually obtainable; whether a needed variable exists in the public release; whether the exposure is a native variable or a construct fitted in the paper; whether the reported number is an adjusted main effect or a pooled/marginal/subgroup quantity; whether the sample is large enough to support the claim. List these in unknowns.
Only OBSERVED failures produce bad.

# Gates, in order

0. DATE. If public_before_cutoff is true or earliest_public_date < 2026-07-01: date_eligible=false, verdict bad, stop. If earliest_public_date is exactly a January 1st, treat it as a placeholder: cap at maybe, unknowns += "date placeholder".

1. SCOPE. Human health, biomedicine, epidemiology, health services, pharmacovigilance, clinical genomics. Physics, ecology, geoscience, materials, economics without a health outcome, bibliometrics about publishing: in_scope_biomedical=false, verdict bad.

2. DATA REALITY (data_feasibility).
   5 open, versioned, downloadable: NHANES/NHIS/BRFSS/MEPS/NSDUH/YRBS/NSCH cycles; CDC WONDER (ICD codes, years, standard population); NHANES public-use Linked Mortality Files; FAERS/VAERS/MAUDE/JADER quarterly files; GEO GSE; TCGA/GDC project + release; GTEx release; GWAS Catalog GCST id, FinnGen release number, IEU OpenGWAS id; ClinicalTrials.gov/AACT snapshot with date; GBD results tool with release; open non-US equivalents (Canadian CCHS PUMF, ENSANUT, Brazil PNS/DATASUS, Peru ENDES, Japan e-Stat/NDB open, KOSIS, Eurostat, WHO Mortality DB).
   4 free registration, standard, versioned: SEER research data (submission year), MIMIC-IV vX.Y, eICU v2.0, KNHANES (KDCA portal), MICS, HRS/ELSA/SHARE/CLSA/CHARLS/CLHLS with waves stated, UK HSE/UKHLS via UKDS.
   3 named source but cycle/version/accession not in the abstract; or DHS (application, usually approved in days); or PRAMS (proposal). Cap at maybe; needs_full_text += "version/cycles".
   2 or less, bad: restricted or licensed: SEER-Medicare, UK Biobank/All of Us individual-level, dbGaP controlled, VigiBase, TriNetX/Optum/MarketScan/national claims, NHIRD, CPRD, ELSA-Brasil, HILDA, HRS sensitive/DNAm; unnamed sources ("a nationwide database", "publicly available GWAS" with no id); live portals without a snapshot (NCBI Pathogen Detection, IMG/M, "all GEO series matching a query", FishBase); data the authors compiled or scraped themselves; "available on request".
   Variables must exist in the mountable release: public SEER has months not days to treatment and no county below state; NHANES has no geocodes; BRFSS no county; MIMIC free text only in the note module. If the estimand needs a missing field: bad, design_flags += variable_not_in_public_release.
   Multi-source: record-level linkage, geocoding of individuals, or harmonising individual-level variables across two or more named cohorts = bad (multi_source_linkage). Joining published aggregate tables on a public key (county FIPS, country-year) from open versioned sources = cap maybe (aggregate_table_join). MR grades the summary statistics, not the underlying cohort (UK Biobank summary statistics are open).

3. ESTIMAND DEFINABILITY (controllability). Would two competent analysts with the same file converge on direction and significance?
   Allowed exposures/outcomes: native codebook variables; standard code lists (ICD, MedDRA PT or SOC, SEER site/histology); closed-form published formulas used continuously or per SD (BMI, BRI, TyG, TyG-BMI, AIP, SHR, UHR, METS-VF, NLR, SII, RAR, eGFR CKD-EPI, EASIX, SHARE-FI, Langa-Weir, KDIGO stages); a gene symbol on a named GSE/TCGA project; a drug name and event PT in FAERS.
   Medium (controllability 3, note as composite_index): literature-defined composite scores with implementation choices: DII/E-DII, CDAI, OBS, HEI-2015, LE8/LC9, MIND/Mediterranean scores, frailty indices, weighted quantile sum mixtures.
   Low (controllability 2 or less, usually bad): exposure or outcome produced by a model fitted in the same paper (trajectory groups, latent classes, k-means or SOM clusters, NLP labels, PCA/factor scores); cut-points that are data-driven, X-tile, "optimal", or unstated ("high vs low expression", "elevated"); undefined ICU phenotypes (sepsis without Sepsis-3 mention, delirium from chart strings, procedure cohorts by string matching, first/peak lab in an unstated window); outcome composites the authors defined (undefined_icu_phenotype or unstated_or_datadriven_cutpoint).
   Estimators allowed: design-weighted prevalence/proportion, logistic/linear/Poisson/modified Poisson/negative binomial/Cox/Fine-Gray, GLM gamma or two-part cost models, direct age standardisation, joinpoint/AAPC, disproportionality ROR/PRR/IC on 2x2 tables, IVW two-sample MR with default clumping, limma/DESeq2 differential expression, Kaplan-Meier differences, adjusted predicted-prevalence contrasts, external validation of a frozen published score (PREVENT, PCE, SCORE2, APACHE, SOFA) reported as AUC or calibration (external_validation_frozen_score, claim_kind auc).
   Estimators that make the headline bad (causal_machinery, psm_or_matching_only, ml_headline): mediation/PROCESS/SEM direct or indirect effects, instrumental variables, staggered DiD, parametric g-formula, age-period-cohort, spatial Durbin, Oaxaca-Blinder, multi-group ITS, clone-censor-weight emulation, ML/SHAP/nomogram performance, a matched or PSM-dependent estimate as the ONLY headline. A simple two-group pre/post or single-series ITS null with the design fully stated may be maybe.
   Headline must be the main effect in the overall population: an interaction, RERI, sex/race-stratified subgroup, one cell of a multi-arm result, "mixed" direction, or the top hit of a screen (many exposures/outcomes/taxa/proteins) is bad unless the screen hit is study-wide significant and the sources are pinned (interaction_or_subgroup_headline, screen_winner).
   A proportion pooled across several surveys or countries by meta-analysis is not a weighted prevalence: controllability 3 at most (pooled_across_surveys). Pooling NHIS across the 2019 redesign, BRFSS across 2011, NSDUH across 2022 needs a stated harmonisation (survey_redesign_pooling).

4. ROBUSTNESS. Use the ARITHMETIC CHECK; do not recompute. robust_margin true, or p <= 0.005, or a powered null (ratio CI within about 0.7-1.4 with analytic N >= 1,000 and >= 100 exposed/events): robustness 4-5. borderline (implied 1.96 <= z < 2.8, or 0.01 <= p < 0.05, or a CI bound within 0.05 of null on the log scale): robustness 2, cap maybe, bad if N < 5,000 or the abstract mentions attenuation in sensitivity analyses (borderline_significance). A null from a tiny or selected sample is a power artefact: robustness 2, cap maybe (underpowered_null). point_outside_ci, asymmetric_ci, p_ci_mismatch, or two different Ns in the abstract: bad (arithmetic_inconsistent). If the abstract gives direction and significance for an allowed estimand but no CI, that is not a failure of the estimand: cap maybe with needs_full_text += "CI". If no number exists anywhere and none could exist, bad. Analytic N < 1,000 or < 100 events for regression items: robustness at most 3.

5. RECALL RISK. Two separate tests; take the worse.
   The BLIND GUESS is your main evidence; the scoring rule is: Track A passes on direction + significance, with CI overlap as the finer score.
   HARD RULE: if the BLIND GUESS direction AND significance both match the paper's primary claim (for a directional claim), recall_risk is at least 4 and the verdict is at most maybe. No exception for "the magnitude is informative": the benchmark's no-data baseline would already pass on direction + significance. Use this rule before any other recall reasoning.
   recall_risk 5: the blind guess matches the paper's direction and significance AND its point guess lies inside the paper's CI (or the claim has no numeric target beyond direction), or known_finding = yes. The no-data baseline would pass: cap at maybe.
   recall_risk 4: the blind guess matches direction and significance but its point guess is outside a narrow CI; or a textbook pair (adiposity and diabetes/ED/asthma, frailty and mortality, smoking and menopause, biomarker ratio and ICU mortality, gene X and prognosis in TCGA, boxed-warning drug and its labelled event, surgery or radiotherapy versus none in SEER, education and fertility; design_flags += textbook_direction) even when the blind guess happened to miss. Verdict at most maybe (replication bucket).
   recall_risk 3: direction plausible but the blind guess got significance wrong, or its numeric guess is far outside the CI, or the estimand is a stratum- or period-specific number.
   recall_risk 1-2: the blind guess is wrong, or says an effect exists where the paper found a null, or the pair is genuinely odd.
   Numeric test for prevalences, rates, counts, AUCs and ROR magnitudes: compare the BLIND GUESS point_guess with the paper's CI: inside means recall_risk >= 4. A wide CI is easier to hit, so wide-CI descriptive numbers are MORE recall-prone. Figures printed in the data producer's own pre-cutoff report (DHS final report, MICS report, SAMHSA tables, NCHS data briefs, KDCA reports) are recallable: recall_risk >= 4.
   Also: if the Background cites a prior study with the same exposure-outcome direction ("prior research using ... found", "we extend"), design_flags += replication_of_cited_prior and recall_risk >= 4. If question_neutral contains a data-derived class name, an analyst-chosen cut-point, or a directional adjective ("delay", "deficiency", "elevated") applied to the exposure, design_flags += question_leaks_result (the question would have to be rewritten; note it, do not reject for it).
   Low recall risk (1-2): null or counter-prior findings, odd exposure-outcome pairs, stratum-specific numeric estimands (HCV RNA prevalence among MSM in NHANES 2011-2014), a specific drug-event ROR, a specific site-period incidence trend, a differential-expression set for one GSE.
   recall_risk 5 means the environment measures memory: cap at maybe (replication bucket).

6. SCORABILITY AND BETTER PRIMARIES. The neutral question must name one estimand; "for each outcome" or several arms means not single. List in your head every scorable claim in the abstract (weighted prevalences with CI, main effects, nulls). If the screener's primary is a poor choice but the abstract contains a better one (a native-variable main effect, a powered null, a stratum-specific numeric estimand), write it in recommended_primary with its value; otherwise write "screener primary". Only promote a null that meets the powered-null bar.

# Verdict

good = "build it": data_feasibility >= 4 (or 3 only because the version is a routine check), controllability >= 4, robustness >= 4 (or a powered null), recall_risk <= 3, the recommended primary is scorable, and no DECISIVE unknown. Routine needs_full_text items never block good.
bad if any gate fails on OBSERVED evidence: date, scope, data_feasibility <= 2, controllability <= 2, arithmetic_inconsistent, headline is a subgroup/interaction/screen winner, no scorable number possible.
maybe otherwise: name the single decisive check first in reasoning. A maybe must always carry at least one decisive unknown, a design flag, recall_risk >= 4, or robustness <= 3; if none applies, the verdict is good.

Family priors from a 320-item human review (stratified, not shortlist rates): registry mortality trends via WONDER 3/4 good; single-survey weighted prevalence of a native variable 7/17; FAERS ROR with quarters 4/15; SEER code-defined cohorts 4/16; single-gene TCGA Cox with continuous exposure 4/26; survey logistic regression 8/55 (mostly textbook pairs); pinned two-sample MR 2/39 (unnamed GWAS and borderline p); ICU cohorts 1/24 (undefined phenotypes, ratio-to-mortality genre); ageing panels 0/29 (home-made indices, waves unstated); latent-exposure or causal-machinery designs 1/48. Do not apply these as quotas; use them to know where to look for the failure.

Prefer variety for the benchmark: nulls, odd pairs, numeric estimands, and non-US families alongside standard survey regressions. Reward clean reproducible measurement, not clinical importance.

# Calibration anchors

- NHANES 2011-2014, MSM aged 20-59, HCV RNA prevalence 3.5% (95% CI 1.9-5.7): open, native variables, weighted proportion, stratum-specific number nobody recalls: good (5/5/5, recall 1).
- SEER 2000-2023 mucinous colorectal cancer, left vs right sidedness, cancer-specific mortality HR 1.33 (1.28-1.39), N=40k, covariates listed, counter-prior direction: good; the submission year and AJCC harmonisation are routine needs_full_text, not blockers.
- CDC WONDER 1999-2024, ICD-defined VTE deaths in adults with cancer, AAMR near doubling, p<0.001: good; the underlying-vs-any-position rule and the AAPC CI are routine full-text items.
- FAERS quarters stated, VOWST and death ROR 0.29 (0.15-0.53), counter-intuitive: good; reference-group and product-name matching rules are routine.
- FinnGen R12 + GCST90205183, IVW MR, OR 0.99 (0.97-1.01) null with 248 instruments: pinned, standard, powered null: good.
- CDC WONDER 1999-2023, ICD J69.0 underlying with I60-I69 contributing, age 25+, AAPC with CI: good.
- NHANES 2005-2018, DII and LE8 < 80, OR 1.39 (1.30-1.50): composite exposure AND composite outcome, textbook direction: maybe (composite_index, textbook_direction).
- MIMIC-IV, AIP quartiles and in-hospital mortality in sepsis, univariable OR only: bad (undefined phenotype, no adjusted estimate, ratio-to-mortality genre).
- CHARLS 2011-2020 staggered DiD on clean heating, "3.3%" with no CI: bad (causal_machinery, no CI).
- VigiBase, any estimand: bad (restricted).
- 2,260 ICH patients, LCMM glucose trajectory group vs 28-day mortality HR 1.71: bad (latent_or_fitted_exposure).
- NSDUH 2022-2023 pooled, prescription opioid use and medical cannabis use, aOR 1.76 (1.53-2.03), native items, post-redesign years: maybe (textbook_direction; precise magnitude is the only non-recallable part).

Output every field of the schema. reasoning <= 80 words.
\end{lstlisting}

\section{Tasks}\label{app:items}
{\scriptsize\setlength{\tabcolsep}{2pt}\renewcommand{\arraystretch}{1.0}%
\newcolumntype{P}[1]{>{\raggedright\arraybackslash}p{\dimexpr #1*92/100\relax}}\begin{longtable}{@{}P{1.55cm}P{4.05cm}P{3.35cm}P{1.85cm}P{1.8cm}P{1.6cm}@{}}
\caption{The 25 benchmark items of window v2026Q3. First public = earliest of preprint, online-first or print date. Kind: D = directional (direction and significance), N = numeric estimate or count, T = trend.}\label{tab:items}\\
\toprule PMID & Paper title & Dataset & Access & First public & Kind / cond. \\ \midrule \endfirsthead
\toprule PMID & Paper title & Dataset & Access & First public & Kind / cond. \\ \midrule \endhead
\bottomrule \endfoot
42505895 & Predicted Excess Cardiovascular Age and a Reverse Socioeconomic Gradient in a Middle-Income Latin American Country: A Population-Based Analysis of 163,889 Peruvians & ENDES Peru & open & 2026-07-09 & D / A \\
42681608 & Racial discrimination in Latin American school settings: evidence from Brazilian adolescents. & PeNSE - Pesquisa Nacional de Saude do Escolar 20 & open & 2026-09-01 & N / A \\
42738955 & Vegetable and Fruit Consumption Among Adolescents in the Gulf Cooperation Council Countries: A Pooled Analysis of National School Surveys & WHO Global School-based Student Health Survey & open & 2026-08-26 & N / A \\
42788097 & Prevalence of night work in the Brazilian employed population. & PNS - Pesquisa Nacional de Saude & open & 2026-09-22 & N / A \\
42621321 & Autoimmune hypothyroidism and endometriosis: evidence from bidirectional Mendelian randomization and an independent clinical cohort analysis & GWAS summary statistics: FinnGen R12-UK Biobank  & open & 2026-08-04 & D / A \\
42700087 & Alzheimer disease and lichen planus: A Mendelian randomization analysis. & GWAS summary statistics & open & 2026-09-01 & D / A \\
42650068 & SPAG1 Expression as a Candidate Predictor of Pathological Lymph Node Metastasis in Prostate Cancer: A Transcriptomic Analysis of The Cancer Genome Atlas Prostate Adenocarcinoma Cohort & TCGA-PRAD & open & 2026-07-28 & D / A \\
42426766 & Who measures the measurers: Closing the administrative accountability gap in non-profit acute care hospitals. & CMS HCRIS & open & 2026-07-09 & N / A \\
42443048 & Substance use among fatally-injured pedestrians in New York State, USA: a retrospective analysis. & NHTSA Fatality Analysis Reporting System & open & 2026-07-13 & N / A \\
42547587 & Real-world safety profile of fezolinetant: a disproportionality analysis based on the FAERS database & FAERS & open & 2026-08-03 & D / A \\
42675671 & A retrospective study of disproportionality analysis signals between proprotein convertase subtilisin/kexin 9 inhibitor use and respiratory, thoracic, and mediastinal disorders. & FAERS & open & 2026-08-28 & D / A \\
42675701 & Safety analysis of acyclovir use in neonates based on the FDA Adverse Event Reporting System: A retrospective observational pharmacovigilance study & FAERS & open & 2026-08-28 & D / A \\
42401752 & Patient out-of-pocket costs for antihypertensive single pill combination products with cost caps & Medical Expenditure Panel Survey, Household Comp & open & 2026-07-04 & N / A \\
42415254 & Treatment Patterns and Barriers to Care Among U.S. Adults With Co-Occurring Substance Use Disorder and Mental Illness & NSDUH & open & 2026-07-08 & N / A \\
42432888 & Cross-sectional association between relative fat mass and abdominal aortic calcification among US adults: The NHANES 2013 to 2014 & NHANES & open & 2026-07-01 & D / A \\
42560253 & Screening for Advanced Liver Fibrosis: Comparing Serum Indices With Elastography in a Nationally Representative U.S. Cohort & NHANES & open & 2026-08-01 & N / A \\
42567448 & A cross-sectional evaluation of racial and ethnic differences in apparent treatment resistant hypertension among US adults & NHANES & open & 2026-08-07 & N / A \\
42578523 & Prevalence of Moderate to Severe Joint Pain Among Adults With Diagnosed Arthritis: United States, 2023 & NHIS & open & 2026-08-11 & N / A \\
42606884 & Serologic Immunity to Hepatitis A and B in US Adults and High-Risk Populations & NHANES & open & 2026-08-17 & N / A \\
42606983 & Prevalence of Urinary Incontinence in the United States 2015-2020 & NHANES & open & 2026-08-17 & N / A \\
42649615 & Prevalence and risk factors of current hepatitis C infection among men who have sex with men aged 20-59 years: NHANES 1999-2016 & NHANES & open & 2026-08-27 & N / A \\
42731386 & Association between bisphenol A exposure, methylmalonic acid, and cardiovascular disease among adults aged 60 and younger & NHANES & open & 2026-09-12 & D / A \\
42736760 & Serum magnesium and depressive symptoms among U.S. adults: A cross-sectional study of NHANES august 2021-august 2023 & NHANES & open & 2026-09-01 & D / A \\
42560347 & A Two-Decade Analysis of Post-Procedural Respiratory Disorder Mortality Among Older Adults in the United States & CDC WONDER Multiple Cause of Death & open & 2026-08-06 & D / A \\
42740601 & Trends and Disparities in Aspiration Pneumonitis Mortality With Co-Listed Stroke Among US Adults, 1999-2024: A CDC WONDER Analysis & CDC WONDER Multiple Cause of Death & open & 2026-09-01 & D / A \\
\end{longtable}
}

\end{document}